\documentclass[10pt,twocolumn,letterpaper]{article}

\usepackage[pagenumbers]{cvpr} %

\definecolor{cvprblue}{rgb}{0.21,0.49,0.74}
\usepackage[pagebackref,breaklinks,colorlinks,citecolor=cvprblue]{hyperref}
\usepackage{array}
\usepackage{float}
\usepackage{listings}
\usepackage{protolang}  %
\usepackage{protostyle} %
\newcommand{\protofield}[1]{{\ttfamily #1}}
\newcommand{\protofieldtype}[1]{\textcolor{blue}{\ttfamily #1}}

\def\paperID{*****} %
\def\confName{CVPR Workshop\xspace}
\def\confYear{2025\xspace}

\title{Synthesizing 3D Abstractions by\\ Inverting Procedural Buildings with Transformers}

\author{\
Maximilian Dax,$^{1,\ast}$ Jordi Berbel,$^2$ Jan Stria,$^3$ Leonidas Guibas,$^2$ Urs Bergmann$^3$\\
{\small
$^1$Max Planck Institute for Intelligent Systems, T\"ubingen, Germany, 
$^2$Google DeepMind,
$^3$Google
}\\
{\small 
$^\ast$Work done at Google. Correspondence: maximilian.dax@tuebingen.mpg.de, ursbergmann@google.com}
}

\begin{document}
\maketitle

\begin{abstract}
  We generate abstractions of buildings, reflecting the essential aspects of their geometry and structure, by learning to invert procedural models. We first build a dataset of abstract procedural building models paired with simulated point clouds and then learn the inverse mapping through a transformer. Given a point cloud, the trained transformer then infers the corresponding abstracted building in terms of a programmatic language description. This approach leverages expressive procedural models developed for gaming and animation, and thereby retains desirable properties such as efficient rendering of the inferred abstractions and strong priors for regularity and symmetry. Our approach achieves good reconstruction accuracy in terms of geometry and structure, as well as structurally consistent inpainting.
\end{abstract}

\section{Introduction}
Abstract visual representations aim to capture the key geometric and structural properties of an object. Abstractions are useful in many applications as they facilitate perceptual comparisons and understanding. For buildings, use cases for abstract representations range from 3D mapping for navigation to the generation of synthetic environments for training of deep learning agents. However, inferring abstractions based on sensor data at a desired level of detail is a challenging problem. Traditional approaches are largely based on geometric simplification and optimization, while learning approaches face a lack of training data.

\begin{figure*}[t]
    \begin{subfigure}[c]{0.25\textwidth}
    \includegraphics[width=\textwidth,trim={0 18.4cm 53cm 0},clip]{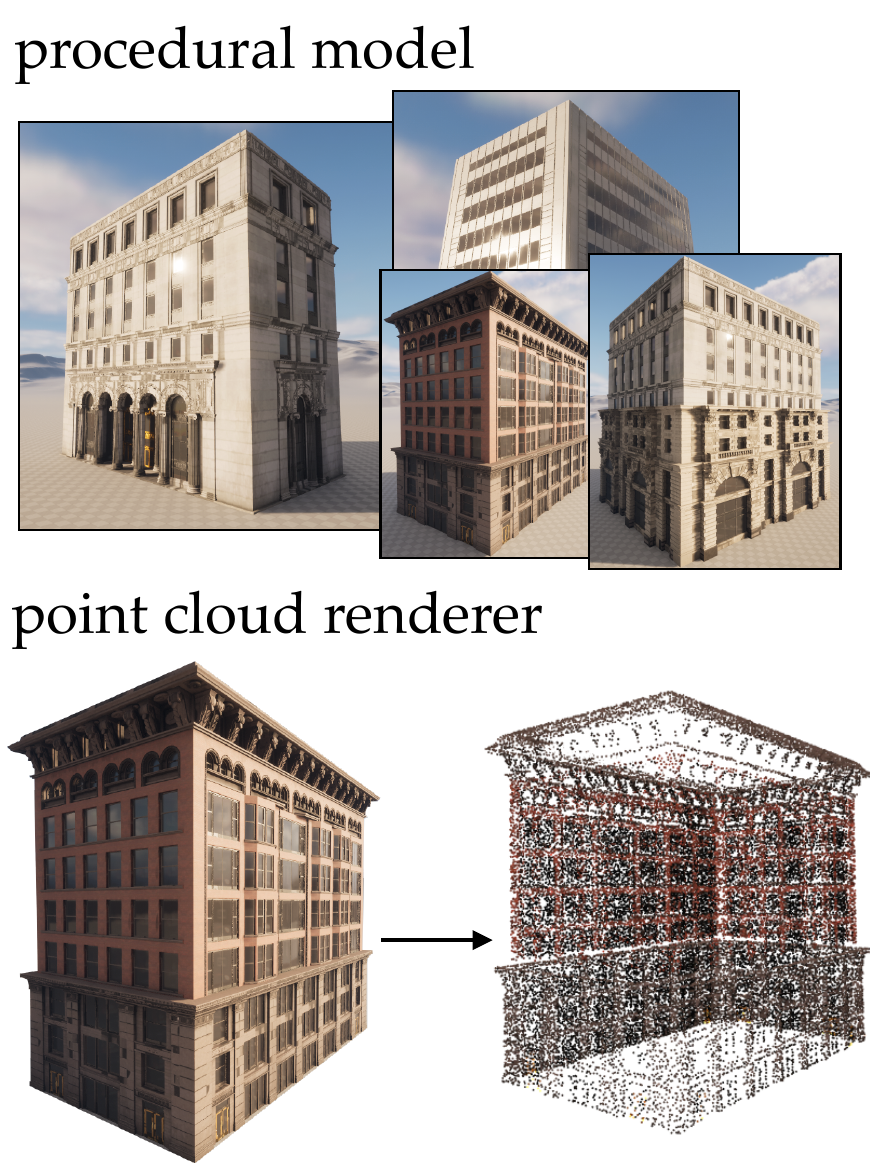}
    \subcaption{Dataset generation}
    \end{subfigure}
    \hfill
    \begin{subfigure}[c]{0.35\textwidth}
    \includegraphics[width=\textwidth,trim={22.5cm 18.4cm 24.5cm 0},clip]{Figures/overview_v2.pdf}
    \subcaption{Inference model}
    \end{subfigure}
    \hfill
    \begin{subfigure}[c]{0.25\textwidth}
    \includegraphics[width=\textwidth,trim={53cm 18.4cm 0cm 0},clip]{Figures/overview_v2.pdf}
    \subcaption{Inference result}
    \end{subfigure}
    \caption{(a) We generate a synthetic training dataset by first unconditionally sampling building abstractions with a procedural model and then composing corresponding point clouds for each abstraction. (b) The inference model encodes the input with a point cloud transformer operating on point cloud voxels (Sec.~\ref{subsec:transformer}). It then employs a language transformer to predict the corresponding abstraction in terms of our custom programmatic language (Sec.~\ref{subsec:dataset}). (c) The transformer output is parsed as a Protocol Buffer and can be rendered in Unreal Engine 5.\looseness=-1
    }
    \label{fig:overview}
\end{figure*}

We here propose a model to infer abstractions from point cloud data. While manually labeled large-scale datasets for this task are costly and thus unavailable, there is an abundance of high-quality 3D building simulators. Specifically, the gaming and animation industries have developed models to sample synthetic environments and render corresponding 3D representations at various levels of complexity---the inverse direction of what we aim to achieve (Fig.~\ref{fig:overview}). Our framework inverts such 3D models with a transformer model~\cite{vaswani2017attention} which takes point clouds as input and predicts corresponding abstractions (Sec.~\ref{sec:method}). Training is fully supervised, based on a dataset of procedural buildings paired with corresponding point cloud simulations. We develop various technical components tailored to the generation of abstractions. This includes the design of a programmatic language to efficiently represent abstractions, its combination with a technique to guarantee transformer outputs consistent with the structure imposed by this language, and an encoder-decoder architecture for the inference model.

Our approach achieves accurate in-distribution geometric and structural reconstruction, and structurally consistent inference for incomplete inputs (Sec.~\ref{sec:experiments}). We find that the main limitations are attributed to constraints of the procedural model, not the inference framework itself. While this can be partially mitigated by data augmentation, our results suggest that procedural model advancements will be crucial for real-world applicability. Based on our analysis, we suggest to make procedural models more flexible to better suit inversion (Sec.~\ref{sec:discussion}).

\textbf{Related Work.} There is vast literature on 3D reconstruction or abstraction from sensor data. A traditional approach to abstraction is geometry simplification---reducing the number of geometric elements used to represent the model~\cite{quadraticsimp,progressivemeshes}. Other works perform geometry abstraction by fitting geometric primitives (e.g., boxes or cylinders) to raw data (e.g., point clouds)~\cite{SPFN,CPFN,shapecoder,inverseCSG,CSG_SGP}. Several works aim to detect symmetries and regularities~\cite{symmetry,regularity,lievoting} or generate structure preserving abstractions~\cite{arrangementnet,lowpolymeshgeneration,niloyabstraction,CLIPasso}. Finally, generative AI has been explored to create geometry from various inputs~\cite{housediffusion,plocharski2024neuro,avetisyan2024scenescript,urbaNprocedural}. In particular, several works represent CAD models and procedural descriptions with a structured, domain specific language---and aim to use language tools for CAD generation~\cite{CADlanguage,sketchgen,vitruvion}.

\section{Method}
\label{sec:method}
We train an inference model that takes a building point cloud $x$ as input and infers a parametric description $\theta$, an abstraction, of the corresponding building. When training the model, we aim to combine \emph{structural knowledge} (e.g. a building is made of storeys/floors, style and position of windows, etc.) and \emph{geometric fit}.

We formalize these notions in a Bayesian framework, and represent \emph{structural knowledge} in the \emph{prior} distribution $p(\theta)$. We further represent the \emph{geometric fit} in terms of the distribution over point clouds $x$ conditional on abstractions $\theta$---the \emph{likelihood} $p(x|\theta)$. With these definitions, the Bayesian posterior $p(\theta|x) \propto p(\theta)p(x|\theta)$ represents the distribution over abstractions $\theta$ for a given point cloud $x$. We solve this Bayesian inference problem by training an inference model $q(\theta|x)$ to approximate the posterior $p(\theta|x)$. An abstraction for a point cloud $x$ can then be inferred by sampling from the trained model $\theta\sim q(\theta|x)$.

Below, we define $p(\theta)$ in terms of a procedural building model and $p(x|\theta)$ in terms of a renderer (Sec.~\ref{subsec:dataset}), introduce the training objective for optimization of $q(\theta|x)$ (Sec.~\ref{subsec:sbi}) and describe the architecture of $q(\theta|x)$ (Sec.~\ref{subsec:transformer}).

\subsection{Synthetic training data}
\label{subsec:dataset}

\textbf{Prior.} 
We employ a procedural building model adapted from the Unreal Engine 5 \emph{City Samples}~\cite{matrixdemo} demo. This model samples abstract buildings $\theta$ by placing a collection of building assets (e.g., facade segments) according to a set of rules (appendix Fig.~\ref{fig:procedural-model}), and hence serves as our implicit prior distribution over buildings $p(\theta)$. Both, assets and rules, have been carefully designed to capture styles of buildings of San Francisco, Chicago and New York. In total there are 911 unique assets, and a typical building is composed of around $10-40$ different asset types. We \textit{represent abstractions} in a custom data format based on Protocol Buffers~\cite{varda2008google} (Fig.~\ref{fig:building_proto_short}), which we designed to facilitate representation of regularities and significantly minimize redundancy, thereby simplifying inference. It is chosen as a hierarchical format, starting from macroscopic structure (height, footprints), and then goes into more detail (storeys, facade description etc). Importantly, identical structures can refer to the same description -- e.g. different storeys can refer to the same facade. For more details, see appendix~\ref{app:technical-details}.

\textbf{Likelihood.}
The likelihood $p(x|\theta)$ models the distribution over point clouds consistent with a given abstraction $\theta$. We represent the likelihood implicitly: for a given abstraction $\theta$ we sample $x\sim p(x|\theta)$ by composing surface points from all assets of a building into a joint point cloud, filtering interior points, and adding Gaussian noise with mean $0$ and variance $\sigma^2$ to each point. As log-likelihood maximization with Gaussian kernels corresponds to mean squared error minimization, this likelihood approximates a geometric distance between point cloud and inferred geometry.

\textbf{Dataset.} We generate a dataset with 341721 pairs of abstractions and point clouds, hence samples from the joint distribution $(\theta,x)\sim p(\theta,x)$ (see Fig.~\ref{fig:overview}a for examples). We augment the asset colors in the dataset by adding uncorrelated Gaussian noise with a standard deviation of 0.15 in the HSV color system to half of the dataset.

\subsection{Training objective}\label{subsec:sbi}

We train the network $q(\theta|x)$ using simulation-based inference~\cite{cranmer2020frontier} ideas (see appendix~\ref{app:sbi} for details) with the loss
\begin{equation}\label{eq:npe-loss}
    L = \mathbb{E}_{x\sim p(x|\theta),\,\theta\sim p(\theta)}
    \left[-\log q(\theta|x)\right],
\end{equation}
across the dataset of simulated buildings from Sec.~\ref{subsec:dataset}. This objective formally corresponds to minimization of the Kullback–Leibler divergence $D_\text{KL}\left(p(\theta|x)\parallel q(\theta|x)\right)$. With sufficient training data and network capacity, the inference network $q(\theta|x)$ will therefore become an accurate estimator of $p(\theta|x)$~\cite{papamakarios2016fast}. After training, the inference network thus inverts the forward model (i.e., the procedural model and point cloud renderer), and an abstraction for a point cloud $x$ can be generated by sampling $\theta\sim q(\theta|x)$. 

\subsection{Model Architecture}
\label{subsec:transformer}
We now define the density estimator $q(\theta|x)$ which maps a point cloud $x$ to a conditional distribution over abstractions $\theta$. We use a transformer-based encoder-decoder model.

\textbf{Tokenization of Protocol Buffers.} 
To obtain a transformer-compatible data format for the abstraction programs $\theta$ we convert these to sequences of tokens. We follow~\cite{ganin2021computer}, which offers a method to bijectively map any Protocal Buffer to a token sequence. Importantly, at inference time, this allows to mask invalid tokens, ensuring syntactically valid Protocol Buffers. For details see appendix~\ref{app:technical-details}.

\textbf{Encoder.} We subdivide the point cloud into cubes of length $7$ meters, resulting in $n_\text{v}$ smaller point clouds $\{\hat{x}_1,\dots,\hat{x}_{n_\text{v}}\}$. Each voxel point cloud is embedded into a feature space $z_i = g(\hat x_i) \in \mathbb{R}^{512}$, where $g$ is a PointCloudTransformer~\cite{guo2021pct} (see Tab.~\ref{tab:architecture} for hyperparameters) and $\hat x_i$ includes point coordinates and color information. We provide the voxel positions by adding sinusoidal positional embeddings~\cite{mildenhall2021nerf} to the PointCloudTransformer's output.

\textbf{Decoder.} We employ a standard transformer decoder~\cite{vaswani2017attention} (see Tab.~\ref{tab:architecture} for hyperparameters). The decoder is conditioned on the embeddings $z_i$ via cross-attention, which enables spatially global information integration.

\textbf{Training.} We train the encoder-decoder model end-to-end for $10^5$ steps, with batch size $16$, employing AdamW~\cite{loshchilov2017decoupled} with $\beta_1=0.9, \beta_2=0.98$ and a learning rate of $5 \cdot 10^{-4}$ after a linear warm up for $10^3$ steps. We apply dropout on the voxel embeddings with rates varying between $0$ and $0.8$, to force the model to infer missing information based on regularity in the data. We regularize with additive Gaussian noise on the point cloud with standard deviations $\sigma$ between $0~\text{m}$ and $0.5~\text{m}$.

\begin{figure*}[t]
    \centering
    \begin{subfigure}[c]{\columnwidth}
        \begin{tabular}{wl{0.65\columnwidth}  wr{0.2\columnwidth}}
            \toprule
            Accuracy number of storeys~~~~ & 97.9\% \\
            Accuracy number of facades & 97.9\% \\
            Accuracy storeys structure & 95.8\% \\
            Assets: precision & 99.4\% \\ 
            Assets: recall & 98.7\% \\ 
            IoU Material Variations & 94.2\% \\
            L2 HSV Color Distance & 0.083\\   
            \bottomrule
       \end{tabular}
      \subcaption{Structural evaluation}
    \end{subfigure}
    \begin{subfigure}[c]{\columnwidth}
    \includegraphics[width=0.95\columnwidth]{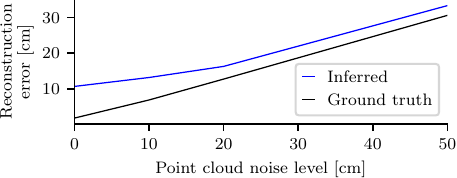}
      \subcaption{Geometric evaluation}
    \end{subfigure}
    \caption{
    Reconstruction performance of the transformer model. 
    (a) Performance on various structural variables (see appendix Tab.~\ref{tab:app-structural-evaluation} for definitions).
    (b) Reconstruction error as a function of the point cloud noise level, measured in terms of the mean deviation between noisy input point clouds and building geometries. The reconstruction error of the inferred buildings (blue) is only slightly larger than the reconstruction error of the ground truth buildings (black). Naturally, both grow with increasing point cloud noise level. 
    }
    \label{fig:reconstruction}
\end{figure*}

\begin{figure*}[t]
    \centering
    \includegraphics[width=\textwidth,trim={0 52.5cm 0 0},clip]{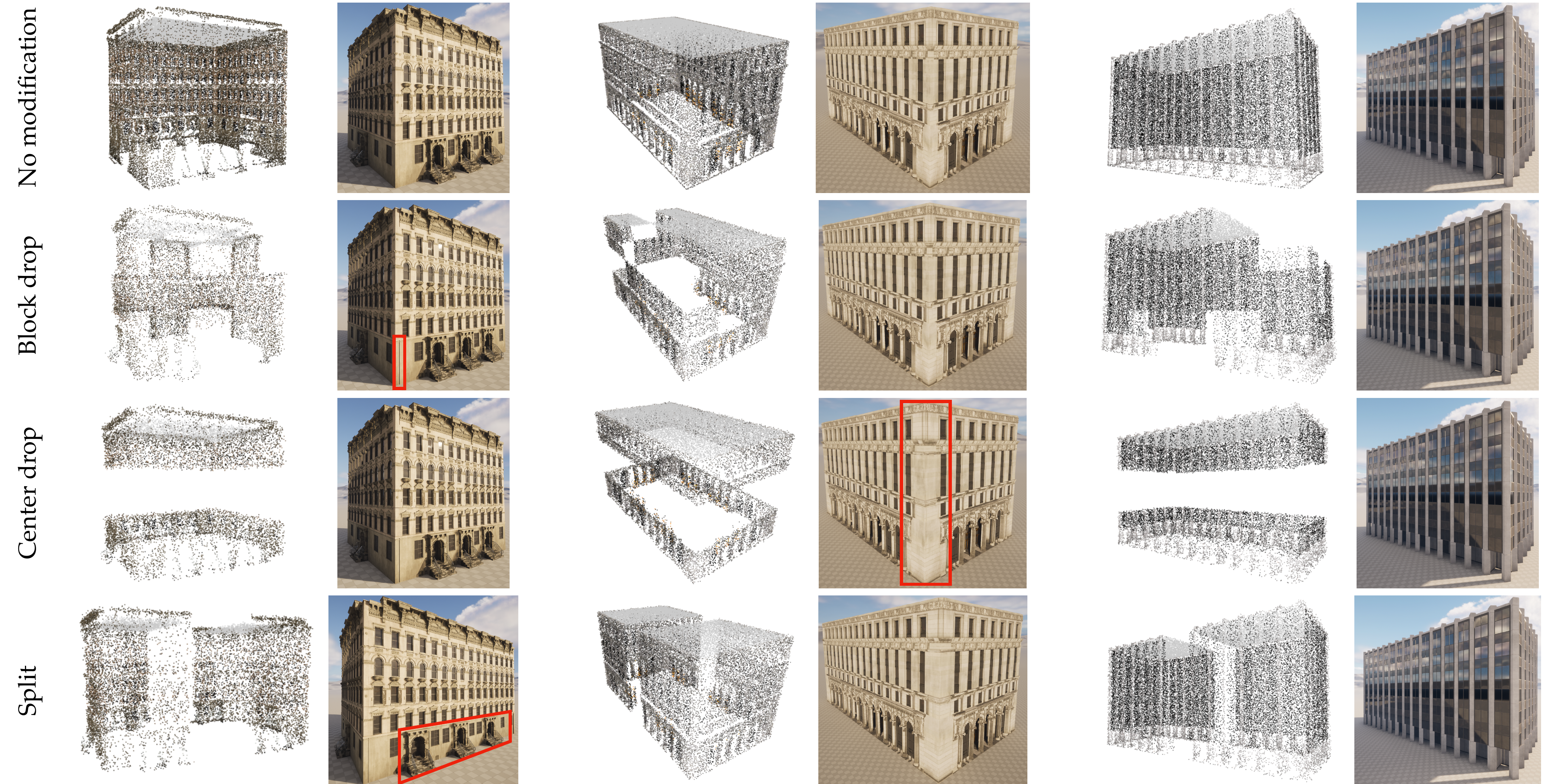}
    \caption{
    Inference results with modified point clouds. We drop random blocks (second row), a single large block in the center (third row), or split the point cloud in the center and move both halves away from each other (fourth row). The inference model successfully reconstructs the associated building. With the split modification, the missing information is filled with additional asset instances, resulting in additional columns of windows. We occasionally observe artefacts (red), like gaps between assets or slight offsets in window positions.
    }
    \label{fig:regularity}
\end{figure*}

\section{Results}\label{sec:experiments}
We first evaluate our trained model on a hold-out subset of $17086$ point clouds from the simulated dataset. For each point cloud, we compare the inferred abstraction to the corresponding ground truth, both in terms of structural properties and geometric fit (Fig.~\ref{fig:reconstruction}). Structural high-level variables, such has the predicted number of storeys of the buildings, the number of facades, and asset precision and recall, are inferred correctly with high accuracies of above $95\%$. As color/materials are augmented for some assets of a building, the intersection over union (IoU) of the predicted and the ground truth modifications is calculated ($94.2\%$), and the Euclidean distance in HSV color space is evaluated on the intersecting assets. The deviation between inferred building geometries and the input point cloud is around 10 cm (in the absence of point cloud noise). We conclude that our model achieves excellent in-distribution reconstruction.

We next test properties of the model under point cloud modifications. 
Specifically, we drop large blocks of the point clouds (a) at random or (b) systematically; and we (c) split the point clouds and move both parts away from each other (Fig.~\ref{fig:regularity}). These manipulations lead to missing information, and (c) additionally could lead to out-of-distribution buildings. We find that the inference model successfully reconstructs missing information by leveraging regularity and symmetry priors from the procedural model. We further observe that inferred buildings for (c) are not simply stretched versions of the originals, but contain additional asset instances (e.g., more windows). While all buildings are visually neat, we observe occasional visual artefacts and slight inaccuracies of asset placements. 

\section{Discussion}\label{sec:discussion}
Our framework casts abstraction into an inverse problem, with the forward direction defined by a procedural model and a point cloud renderer. The inference network is trained to invert this forward direction, thereby converting point cloud data to the procedural model space. In this space, buildings are composed of assets, which can directly serve as abstract representations: they allow for high-level editing and rendering at different levels of detail.

\textbf{Limitations.}
As a central limitation, our approach is constrained to the scope of the forward direction, hence is limited to buildings that are captured by the procedural model. While procedural models generate \emph{diverse} buildings, they usually are not \emph{comprehensive}, because for simulation of realistic environments it is not necessary to capture \emph{all} buildings within the targeted distribution. Therefore, the procedural model used here is limited to only a subset of buildings.
Application to real buildings further requires the point clouds to be consistent with real point cloud measurements (e.g., with photogrammetry, lidar). We designed our renderer to capture some of the main features of realistic point clouds (e.g., only rendering points from the surface, added noise), but more realistic settings require further extension (e.g., local variation of noise levels). Additionally, it may be necessary to explicitly account for domain shifts between simulated and real data within our model~\cite{cannon2022investigating,wehenkel2024addressing}.

\textbf{Strengths.}
Our simulation-based approach has various advantages compared to other methods.
First, it enables training without human annotations with large amounts of synthetic data. Second, regularity priors implemented in the procedural model are captured by the trained inference model. Buildings are highly regular objects, and such priors enable inference of missing information. Third, our approach provides an inductive bias to prioritize structure over geometric accuracy, as it is defined on the abstract descriptions (as opposed to a geometric reconstruction loss). Indeed, we observed that inferred buildings sometimes did not optimally capture the input point cloud, but always looked structurally self-consistent. This is desirable, as structure is typically more important than rigorous geometric accuracy for abstractions. Fourth, the native representation of procedural models is optimized for extremely efficient rendering, which is inherited by our framework. Fifth, our approach enables human-in-the-loop interaction between procedural modeling and its inversion. This could help to quantitatively assess procedural models at scale, validating how well they capture real buildings and identifying improvements.

\textbf{Optimizing procedural models for inversion.}
Designing procedural models with inversion in mind likely improves the scope and performance of our framework. 
E.g., our experiments suggest a need for more flexible asset selection and placement. This could alleviate inconsistencies between asset placement and augmented point clouds (e.g., Fig.~\ref{fig:regularity}, split manipulation). Parametric assets, e.g. variable window sizes, would further enhance flexibility, which however also weakens the regularity prior. Such modifications to procedural models will be crucial for future applicability of our framework to real data.

\clearpage
{
    \small
    \bibliographystyle{ieeenat_fullname}
    \bibliography{main}
}

\clearpage

\appendix
\renewcommand{\thesection}{A.\arabic{section}}
\setcounter{section}{0}

\setcounter{page}{1}
\renewcommand*{\thepage}{A\arabic{page}}
\maketitlesupplementary

\section{Relation to simulation-based inference}\label{app:sbi}

Our framework for procedural inversion is closely related to simulation-based inference~\cite{cranmer2020frontier} (SBI). SBI is a paradigm for solving inverse problems entirely based on \emph{simulations} from the forward model, without requiring access to underlying densities. SBI has become a standard tool for scientific inference~\cite{brehmer2018constraining,gonccalves2020training,Dax:2021tsq,hermans2021towards}, and a variety of methods have been developed in recent years~\cite{papamakarios2016fast,lueckmann2017flexible,greenberg2019automatic,sisson2018overview,beaumont2002approximate,hermans2019likelihood,thomas2020likelihoodfree,miller2022contrastive,papamakarios2019sequential,lueckmann2019likelihood,sharrock2022sequential,geffner2022score,Dax:2022pxd,wildberger2023flow}. Our approach can be seen as a variant of neural posterior estimation~\cite{papamakarios2016fast,lueckmann2017flexible,greenberg2019automatic} (NPE), an SBI method which directly targets the Bayesian posterior.

However, compared to common scientific use cases, our application of SBI to infer abstractions of buildings has various distinct properties. First, our inference domain (i.e., abstractions $\theta$) is represented in terms of a programmatic language while in scientific applications it is typically a low-dimensional vector space. Our density estimator $q(\theta|x)$ is thus parameterized with a transformer model as opposed to a continuous density estimator such as a normalizing flow~\cite{rezende2015variational,papamakarios2019normalizing}. Second, our prior is a complicated procedural model whereas in scientific applications it is typically a simple distribution and the complexity of the problem is dominated by the likelihood. Third, we are ultimately interested in generating individual high-likelihood abstractions, not necessarily in the distributional properties of the posterior such as correlations and uncertainties.

\begin{figure}[t]
    \begin{subfigure}[c]{0.48\columnwidth}
    \includegraphics[width=\textwidth,trim={0 0 0 0cm},clip]{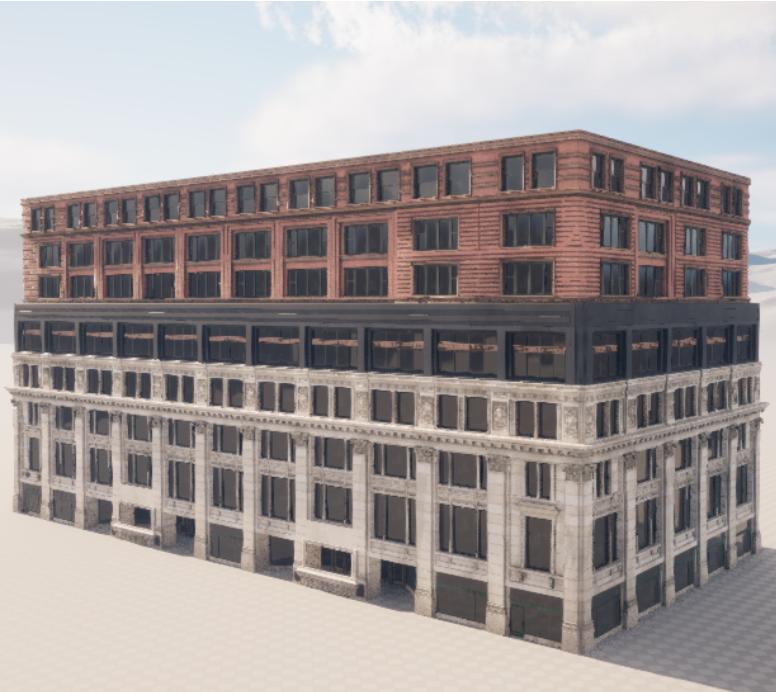}
    \subcaption{Procedural building}
    \end{subfigure}\hfill
    \begin{subfigure}[c]{0.48\columnwidth}
    \includegraphics[width=\textwidth,trim={0 0 0 0cm},clip]{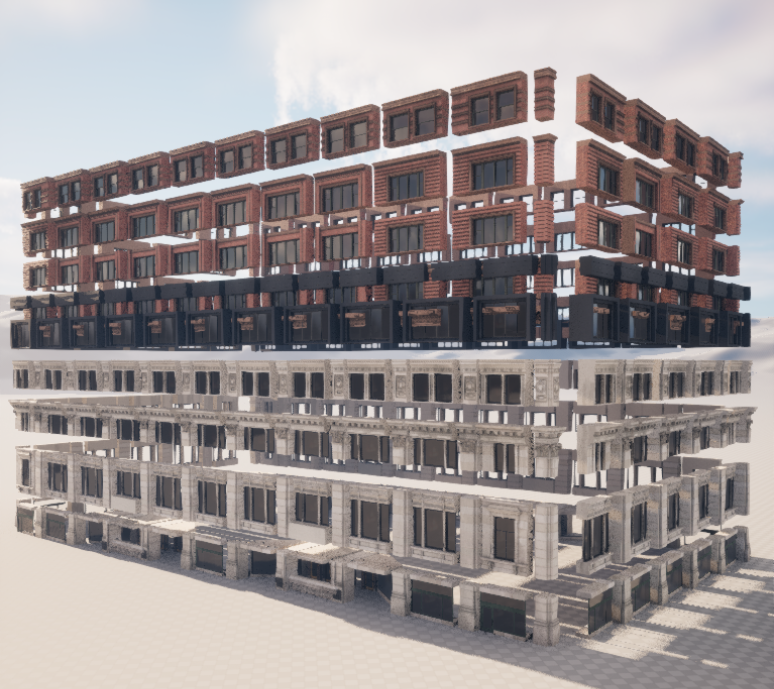}
    \subcaption{Assets in procedural building}
    \end{subfigure}\vspace{0.2cm}
    \caption{
    A procedural building is generated by placing a set of assets according to a set of handcrafted rules.
    }
    \label{fig:procedural-model}
\end{figure}

\begin{table}[t]
    \centering
    \begin{tabular}{l l}
        \toprule
        Description\hspace{2cm} & Protocol Buffer fields\hspace{4.1cm}\\\midrule
        Noise level$^\ast$ & \protofield{noise\_level}\\
        Absolute coordinates$^\ast$ & \protofield{height},\protofield{x}, \protofield{y}, \\
        Relative coordinates$^\ast$ & \protofield{offset}, \protofield{elevation}\\
        Asset scales$^\ast$ \& rotations$^\ast$ & \protofield{scale\_x}, \protofield{scale\_y}, 
        \\ & \protofield{quaternion\_3},
        \\ &  \protofield{quaternion\_4}\\
        Asset index & \protofield{cell\_type}\\
        Pointer index & \protofield{facade\_index}, \\& \protofield{footprint\_index}\\
        \bottomrule
    \end{tabular}
    \caption{Token groups for the fields of the Protocol Buffer representing the programmatic language for abstractions $\theta$ (omitting prefixes, see Fig.~\ref{fig:building_proto_short} and F for complete definitions). We use continuous groups (marked with asterisk) for absolute coordinates of the building, for relative coordinates within a storey and for scale and rotation parameters of assets. We use discrete integer groups for determining asset types from an external asset index and for pointer indices within the Protocol Buffer.}
    \label{tab:token-groups}
\end{table} 

\section{Technical details}\label{app:technical-details}

\textbf{High-level Description of the Protocol Buffers.}
A building is chosen to be represented in a hierarchical description with a Protocol Buffer. This hierarchical format starts with the macroscopic structure in terms of the building \protofield{height} and (possibly multiple) \protofield{footprints} layouts. It then specifies a series of \protofield{storeys}. A storey is defined in terms of its \protofield{elevation} and by linking to a \protofield{facade}. A facade is specified by a sequence of \protofield{cells\_patterns}, which in turn contain collections of asset instances. The nested structure of this data format provides a natural way of representing recurring patterns. For example, multiple storeys can link to the same facade, and multiple facades can link to the same footprint layout. Where appropriate, we define coordinates relative to previously specified data (Tab.~\ref{tab:token-groups}). For example, asset positions are defined in relative terms to the associated footprint line segment, and their horizontal scales are computed as the difference to respective previously placed assets, enforcing consistency between asset placements and the building footprints. To account for (rare) cases, where asset properties differ from the derived choice, we add a custom \protofield{cell\_modifier}. Finally, each asset uses one or more materials, and the building Protocol Buffer contains a set of color modifiers for material and asset pairs. If unset for an asset's material, the standard color of the material is used.

\textbf{Complete definition of the Protocol Buffers.}
Fig.~\ref{fig:building_proto_short} defines the Protocol Buffer we used, and Fig.~\ref{fig:building_proto_full} displays the additional omitted definitions. A Footprint is parameterized as a planar polygon. The line segments of the footprint determine the layout of the corresponding facades, along which the asset cells are arranged.

A CellModifier can be used to change the scale and rotation of a Cell instance (which wraps a single asset instance). In most cases, scales and rotations can be uniquely determined based on heuristic rules (e.g., using the cell position within the footprint and coordinates of neighbouring cells). In our definition of the Protocol Buffer, we thus do not specify cell scales and rotation by default, and instead derive them based on such heuristics. This removes redundancy from the Protocol Buffer representation and also decreases the length of the corresponding tokenized sequences. However, in some edge cases, these rules don't provide a unique result. In such cases, the optional CellModifier can be used to overwrite scale and rotation parameters.

Finally, MaterialVariation optionally modifies the material type and color parameters of a specific asset. When applied, this modifies all asset instances of the specified type in the same way.

\begin{figure*}[t]
    \centering
\begin{lstlisting}[language=protobuf3,style=protobuf]
// A cell wraps a single asset instance, adding a positional parameter and optional modifiers.
message Cell {
  int32 cell_type;                                    // Enumeration into an external index
  float offset;                                       // Offset along the corresponding linesegment
  optional CellModifier cell_modifier;                // Modifier that overwrites scales and roations
}

// A cell pattern combines a sequence of asset cells along one line segment of the footprint
message CellsPattern {
  repeated Cell cells;                                // Sequence of cells
}

// A facade is made by defining the cells along all line-segments of a footprint
message Facade {
  repeated CellsPattern cells_pattern;                // A pattern of asset instances
  int32 footprint_index;                              // Index to the footprint array
}

// A storey is made by combining a footprint layout with a facade pattern
message Storey {
  int32 facade_index;                                 // Sets which facade to use for this storey
  float elevation;                                    // Relative in [0, 1] * Building.height
}

message Building {
  float noise_level;
  float height;                                       // Total height of the building
  repeated Footprint footprints;                      // Footprint polygons
  repeated Storey storeys;                            // Building storeys
  repeated Facade facades;                            // Unique facade patterns
  repeated MaterialVariation material_variations;     // Material variation (colors, material type)
}
\end{lstlisting}
    \caption{Custom format for the representation of abstract buildings. This hierarchically combines asset instances (``Cells'') into recurring patterns (``CellsPattern''). These patterns are combined into facade instances, which can in turn be linked by the building storeys. Finally, a building is composed by combining such storeys along with variables characterizing the high-level geometry (height, footprint polygons) and the point cloud noise level. Definitions of CellModifier, Footprint and MaterialVariation are provided in Fig.~\ref{fig:building_proto_full}.}
    \label{fig:building_proto_short}
\end{figure*}

\begin{figure*}[t]
    \centering
\begin{lstlisting}[language=protobuf3,style=protobuf]
// 2D vector
message Vector2d {
  double x;
  double y;
}

// 2D polygon for building footprints (different building storeys can have different footprints)
message Footprint {
  repeated Vector2d points;
}

// Cell modifier to overwrite default scales and rotations of assets
message CellModifier {
  float scale_x_overwrite;                            // Used only if can't predict.
  float scale_y_overwrite;                            // Used only if can't predict.
  float quaternion_3_overwrite;                       // In case orientation is wrong.
  float quaternion_4_overwrite;                       // In case orientation is wrong.
}

// List of material types for assets
enum MaterialBaseType {
  UNSPECIFIED;
  GLASS;
  BRICK;
  PAINTEDMETAL;
  LIMESTONE;
  PAINTEDSTONE;
  GRANITE;
  WOOD;
  METAL;
  COPPER;
  BRASS0;
  BRUSHED1;
  ASPHALT2;
  DIRTY3;
  ROOFTOP4;
  SIDEWALK5;
  BLOCK6;
}

// Material variation, defining modifications to specific asset types
message MaterialVariation {
  int32 cell_type;                                    // Which asset type to modify
  MaterialBaseType material_base_type;                // Optional change of asset material
  float hue;                                          // Color hue in HSV
  float saturation;                                   // Color saturation in HSV
  float brightness;                                   // Color brightness in HSV
}
\end{lstlisting}
    \caption{Protocol Buffer definition of base objects referred to in Fig.~\ref{fig:building_proto_short}.}
    \label{fig:building_proto_full}
\end{figure*}

\begin{table}[t]
    \centering
    \begin{tabular}{l r r}
        \toprule
        Description\hspace{2cm} & Range & Resolution\\\midrule
        Noise level [m] & $[0, 1]$ & 0.01 \\
        Absolute coordinates [m] & $[-100, 100]$ & 0.1\\ 
        Relative coordinates & $[0, 1]$ & 0.005\\
        Asset scales \& rotations & $[-5, 5]$ & 0.01\\
        \bottomrule
    \end{tabular}
    \caption{
        Discretization of continuous variables for tokenization. Noise level and absolute coordinates are specified in meters.
    }
    \label{tab:token-groups-continuous}
\end{table} 

\textbf{Tokenization of Protocol Buffers.}
\label{app:pb-tokenization}
To obtain a transformer-compatible data format for the abstraction programs $\theta$ we need to convert these to sequences of tokens. Following~\cite{ganin2021computer}, we leverage the known structure of the Protocol Buffers to obtain an efficient conversion scheme. We first define token groups for the different data types in the Protocol Buffer (Tab.~\ref{tab:token-groups}) and assign to each group a set of tokens that covers all potential values. We further use a set of special tokens in the sequence to navigate within the Protocol Buffer whenever the next step is not unique. This includes end tokens for \protofieldtype{repeated} fields and selector tokens for \protofieldtype{optional} and \protofieldtype{oneof} fields (see~\cite{ganin2021computer} for details). 

This scheme can convert any Protocol Buffer representation to a sequence of tokens. Conversely, at inference time we can mask out invalid options for the next token at each step of the autoregressive transformer prediction, such that any inferred sequence can be converted back into the Protocol Buffer representation. This bijective mapping between Protocol Buffers and token sequences enables straightforward application of standard language modeling techniques for the estimation of the programs $\theta$ representing building abstractions.

Conversion of the Protocol Buffers into the tokenized representation requires assignment of discrete tokens to each possible value for each Protocol Buffer field. Following~\cite{ganin2021computer}, we group fields with similar functions together (Tab.~\ref{tab:token-groups}; e.g., \{\protofield{facade\_index},\protofield{footprint\_index}\}, which are both used to cross-reference objects within the Protocol Buffer, or \{\protofield{height}, \protofield{x}, \protofield{y}\}, which all refer to spatial coordinates). Continuous values further need to be discretized (Tab.~\ref{tab:token-groups-continuous}).

\begin{table}[t]
    \centering
        \centering
        \begin{tabular}{l r}
            \toprule
            \multicolumn{2}{c}{Encoder (PointCloudTransformer~\cite{guo2021pct})}\\\midrule
            Voxel size & ~~~~$7~\text{m}\times7~\text{m}\times7~\text{m}$ \\
            Max. \# points per voxel & 300 \\
            \# attention layers & 4 \\
            \# attention heads & 4 \\
            Dimension of QKV & 256 \\
            Output dimension & 512 \\
            \bottomrule
            \\
            \toprule
       \multicolumn{2}{c}{Decoder (Transformer~\cite{vaswani2017attention})}\\\midrule
            Size of context window & ~~~~2048 \\
            \# attention layers & 12 \\
            \# attention heads & 8 \\
            Dimension of QKV & 512 \\
            \bottomrule
        \end{tabular}
    \caption{Hyperparameters of the inference model.}
    \label{tab:architecture}
\end{table} 

\section{Additional results}

\begin{figure}[H]
    \centering
    \includegraphics[width=\columnwidth]{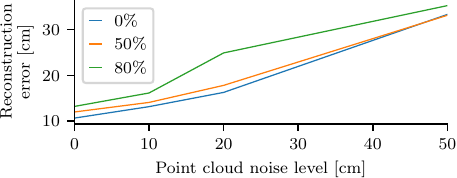}
    \caption{Reconstruction error as a function of the point cloud noise level. Compared to Fig.~\ref{fig:reconstruction}, this also shows results for input point clouds with 50\% and 80\% dropout augmentation of point cloud voxels.}
    \label{fig:reconstruction_dropout}
\end{figure}

Tab.~\ref{tab:app-structural-evaluation} defines the structural evaluation quantities from Tab.~\ref{fig:reconstruction}. Fig.~\ref{fig:reconstruction_dropout} expands over Fig.~\ref{fig:reconstruction}, also showing the reconstruction accuracy for input point clouds with voxel dropout. The reconstruction error is a bit higher in case of dropout, which indicates that the inference model uses global point cloud information, which also improves its local estimates.
\vspace{1cm}

\begin{table*}[]
    \centering
        \begin{tabular}{wl{0.22\textwidth}  wl{0.73\textwidth}}
            \toprule
            Name & Description \\
            \midrule
            Accuracy number of storeys & Fraction of inferred buildings with correct number of storeys. \\
            Accuracy number of facades & Fraction of inferred buildings with correct number of distinct facades. \\
            Accuracy storeys structure & Fraction of storeys which link to the correct facade.\\
            Assets: precision & Fraction of distinct assets in inferred building, which are also present in ground truth building. \\ 
            Assets: recall & Fraction of distinct assets in ground truth building, which are also present in inferred building. \\ 
            IoU Material Variations & Intersection over union of modified materials in inferred and ground truth building. \\
            L2 HSV Color Distance & L2 distance between ground truth and inferred material colors in HSV basis.\\
            \bottomrule
       \end{tabular}
    \caption{Description of metrics used in Fig.~\ref{fig:reconstruction}.}
    \label{tab:app-structural-evaluation}
\end{table*}

\end{document}